\documentclass{article}

    \PassOptionsToPackage{numbers, compress}{natbib}

    \usepackage[preprint]{neurips_2021}

\usepackage[utf8]{inputenc} 
\usepackage[T1]{fontenc}    
\usepackage{hyperref}       
\usepackage{url}            
\usepackage{booktabs}       
\usepackage{amsfonts}       
\usepackage{nicefrac}       
\usepackage{microtype}      
\usepackage{xcolor}         

\usepackage{amsmath}
\usepackage{bm}
\usepackage{graphicx}
\usepackage[ruled]{algorithm2e}
\usepackage{caption}
\usepackage{subcaption}
\usepackage{wrapfig}

\title{Negative Feedback System as Optimizer for Machine Learning Systems}

\author{%
  Md Munir Hasan\thanks{Use footnote for providing further information
    about author (webpage, alternative address)---\emph{not} for acknowledging
    funding agencies.} \\
  Department of Electrical and Computer Engineering\\
  University of North Carolina at Charlotte\\
  Charlotte, NC 28223 \\
  \texttt{mhasan13@uncc.edu} \\
   \And
   Jeremy Holleman \\
   Department of Electrical and Computer Engineering\\
   University of North Carolina at Charlotte\\
   Charlotte, NC 28223 \\
  \texttt{mhasan13@uncc.edu} \\
}

\begin{document}

\maketitle

\begin{abstract}
With high forward gain, a negative feedback system has the ability to perform the inverse of a linear or non-linear function that is in the feedback path. This property of negative feedback systems has been widely used in analog electronic circuits to construct precise closed-loop functions. This paper describes how the function-inverting process of a negative feedback system serves as a physical analogy of the optimization technique in machine learning. We show that this process is able to learn some non-differentiable functions in cases where a gradient descent-based method fails. We also show that the optimization process reduces to gradient descent under the constraint of squared error minimization. We derive the backpropagation technique and other known optimization techniques of deep networks from the properties of negative feedback system independently of the gradient descent method. This analysis provides a novel view of neural network optimization and may provide new insights on open problems.
\end{abstract}

\section{Introduction}
Gradient descent has long been the dominant method for optimizing weights in neural networks. It is constructed purely from a mathematical point of view with the goal to minimize a loss function. Many mathematical formulations are modeled after a physical process. The most relevant example is the deep neural network which is modeled after a biological process. Having a physical process behind a mathematical model has the advantage that the behavior of the physical process can provide intuition for the mathematical model. For example, the convolutional neural network~\cite{Fukushima1980}, which now forms the backbone of image recognition, is inspired by the receptive field of the mammalian visual cortex~\cite{Hubel1968} . Gradient descent with momentum is developed by analogy with stabilizing a heavy ball rolling down a hill. We believe that studying the physical process which describes the optimization should help us design a better optimizer. In this paper we present a negative feedback system as a physical analogy of optimization and show a close relationship to gradient descent. This optimization method is based on the ability of a negative feedback system to perform the inverse operation of a function. This principle is well known in the analog circuits and systems community and many useful analog circuits have been constructed~\cite{Sarpeshkar2009} using this principle. Our contributions in this paper can be summarised as follows.
\begin{itemize}
    \item We develop the method of using negative feedback system as an optimizer for the field of machine learning. We state the assumptions and establish necessary conditions to achieve convergent learning.
    \item We show that some non-differentiable functions that cannot be optimized through gradient descent can be learned using a negative feedback learning rule. We establish the conditions under which it works and provide a physical interpretations of these conditions. 
    \item We show that the negative feedback system reduces to gradient descent under a squared error minimization constraint. We also derive the error backpropagation technique from within a negative feedback system and show that it is the same as backpropagation under gradient descent. 
    \item We show that many optimization techniques such as weight decay, adaptive learning~\cite{Kingma2015AdamAM}, and residual networks~\cite{He2016}, which were previously developed independently of each other, can be described by the framework of a negative feedback system. We also offer a possible solution to the weight transport~\cite{Qilani2016} problem, one of the major barriers to a biologically plausible neural network model.
\end{itemize}

\begin{figure}[!t]
\centering
\subfloat[]{ \includegraphics[width=2.5in]{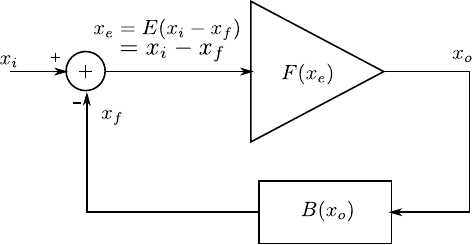} \label{fig:general-fb} } 
\hfill
\subfloat[]{ \includegraphics[width=1.5in]{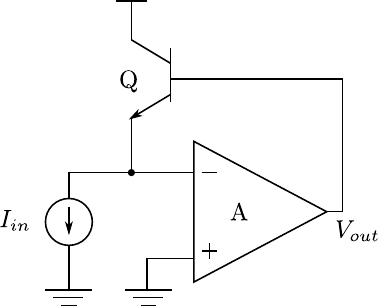} \label{fig:i-v-ckt} }
\caption{(a) A generic negative feedback system (b) An Operational amplifier with an exponential element in the feedback path realizes a logarithmic input-output function. The transistor Q has exponential voltage to current relationship. The feedback system implements inverse of the exponential i.e. logarithmic function.}
\end{figure}

\section{Theoretical Background} \label{sec:principle}
For a negative feedback system as shown in Fig.~\ref{fig:general-fb}, if we define the forward function, backward function, and the error function with Eq.s~(\ref{eq:forward-function}), (\ref{eq:backward-function}) and (\ref{eq:error-function}) respectively, then the input output relationship is expressed by Eq.~(\ref{eq:in-out}). For a forward function of the form $y = F(x) = Ax$ where $A$ is the gain, the inverse of the forward function is $x = F^{-1}(y) = y/A$. If the gain $A$ is large then $F^{-1} \to 0 $. For error function of the form $y = E(x) = ux$ where $u$ is the gain, $E^{-1}(F^{-1}) \to 0 $ for high forward gain $A$. Then the output of the feedback system becomes inverse of the backward function as in Eq.~(\ref{eq:inv}). Effectively, the negative feedback system is implementing the inverse of the function that is in the backward path. 

\begin{align}
    x_o &= F(x_e) \quad \text{(forward function)} \label{eq:forward-function}\\
    x_f &= B(x_o) \quad \text{(backward function)}\label{eq:backward-function}\\
    x_e &= E(x_i - x_f) \quad \text{(error function)}\label{eq:error-function}\\
    F^{-1}(x_o) &= E(x_i - B(x_o)) \label{eq:in-out} \\
    x_i &= E^{-1}(F^{-1}(x_o)) + B(x_o) \\
    x_i &\approx B(x_o) \quad \text{(for large $A$, $E^{-1}(F^{-1}(x_o)) \to 0$})\\
    x_o &= B^{-1}{(x_i)} \label{eq:inv}
\end{align}
    
This property is commonly used in analog circuits in order to perform inverse operation of the transistor function~\cite{Sarpeshkar2009, Liu2002}. An example circuit is shown in Fig.~\ref{fig:i-v-ckt}. In a transistor an input voltage creates an exponential output current. However, the transistor is an uni-directional device which means that pushing a current at the output of the transistor will not produce a voltage at the input. In order to make that operation work, a negative feedback system using an operational amplifier of gain $A$ is used which implements that inverse operation. This way an input current $I_{in}$ into the feedback system produces the corresponding transistor voltage $V_{out}$. 

It should be noted that even if the backward function $B$ is not completely invertible (which is the case for an uni-directional transistor), the overall system appears to be performing $B^{-1}$. This is because the system is not using $x_i$ (the range of $B$) as input to the function $B^{-1}$ directly. Rather, as in Fig.~\ref{fig:general-fb}, the output of the system $x_o$ acts as the domain of $B$. The output of $B$ is then compared with the target range of $B$ i.e. $x_i$. When the difference of $x_i$ and $x_f$ is zero, the overall system output $x_o$ is approximately the output of $B^{-1}$.

\begin{figure}[!t]
\centering
\includegraphics[width=2.5in]{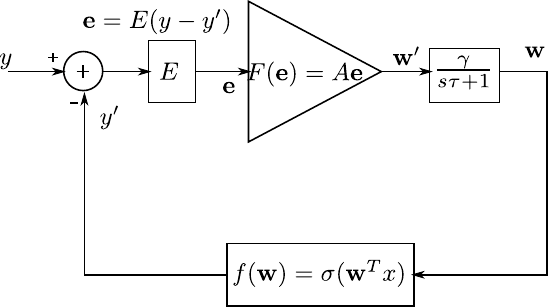} 
\caption{A negative feedback system as optimizer for machine learning system.}
\label{fig:system-setup}
\end{figure}

\section{Method}\label{sec:method}
\subsection{System Setup}
To frame optimization as a negative feedback problem, we express the a layer as a function of the weights, with the inputs held constant.
In a neural network, a single layer can be expressed as a function of a linear combination of $\mathbf{x}$ with a weight vector $\mathbf{w}=[w_1,w_2,\hdots,w_n]^T$ as shown in Eq.~(\ref{eq:formulation-1}). There can be linear or non-linear activation function $\sigma$ inside the function $f$. A bias term can be easily implemented by setting an element of the $\mathbf{x}$ vector to $1$. The variables $x_i$ and $y$ are training samples which are known quantities for a problem. By implementing the inverse operation of the function in Eq.~(\ref{eq:formulation-2}) we can find the weights $w_i$, which effectively implements an optimization operation. 
\begin{align}
    y &= f(\mathbf{w}) = \sigma(\sum_i w_i x_i) \label{eq:formulation-1}\\
    \mathbf{w} &= f^{-1}(y) \label{eq:formulation-2}
\end{align}

To implement the inverse operation using negative feedback, the function $f$ is placed in the feedback path as shown in Fig.~\ref{fig:system-setup}, $x$ training samples are used in the backward function, weights are initialized randomly and $y$ training samples are set as input to the feedback system. An initial prediction of the weight vector $\mathbf{w}$ is used by the backward function to produce $y'$. Using the difference $y-y'$ an error $\mathbf{e}$ is generated. The process of generating a vector $\mathbf{e}$ with a scalar $y-y'$ is described in the following subsection. 

\subsection{Stability Criteria}\label{sec:stability-criteria}
In order for a feedback system to be stable, the bandwidth of the system should be limited, meaning that the output should change slowly (a low frequency system). Hence, instead of changing the weight from the previous value to the new value predicted by the forward function instantly (infinite bandwidth), a small increment is made from the previous value toward the predicted value by using a first order low pass filter as shown below. 
\begin{align}
    \frac{\mathbf{w}}{\mathbf{w'}} &= \frac{\gamma}{s\tau+1} \quad \text{(Laplace transformed low pass filter transfer function)} \nonumber \\
    \tau\frac{\partial \mathbf{w}}{\partial t} &= -\mathbf{w} + \gamma\mathbf{w'} \nonumber \\
    \mathbf{w}^{t} &= \mathbf{w}^{t-1} + (\gamma\mathbf{w'} - \mathbf{w}^{t-1})\frac{\partial t}{\tau} =
    \mathbf{w}^{t-1} + (A\gamma\mathbf{e} - \mathbf{w}^{t-1})\eta \label{eq:w-update-formulation}
\end{align}
This is similar to using  a small learning rate in gradient descent. The prediction labeled $\mathbf{w'}$ from the forward function goes into a low pass filter characterised by a time constant $\tau$ and arbitrary constant $\gamma$ which outputs slowly varying $\mathbf{w}$. This new value of $\mathbf{w}$ goes around the feedback loop again and with consecutive iterations around the feedback loop, the output converges to the optimum value of $\mathbf{w}$. The weight update method because of the low pass filter is given in Eq.~(\ref{eq:w-update-formulation}) where the quantity $\eta = \partial t/\tau$ acts as the learning rate. The superscript $t$ denotes the weight at time $t$ during iteration. 

Another important criterion for stability is that the gain around the feedback loop must be negative when the magnitude is greater than unity~\cite{astroem2021feedback}. From Fig.~\ref{fig:system-setup}, the forward gain is $A$ and the backward gain is $\bm{\beta} = {\partial {y'}}/{\partial \mathbf{w}}$. The loop gain of the system is $-1\times A\bm{\beta}$. Hence, we have to make sure that the product of the forward and backward gain for each component of $\bm{\beta}$ is always positive. The forward gain $A$ is typically positive. If any component of $\bm{\beta}$ is negative for a training sample then the corresponding element of the gain product becomes negative. In general, if we use a forward gain of $A\bm{\beta}$, then the element-wise product of forward and backward gain is $A\bm{\beta}\times\bm{\beta} = A\bm{\beta}^2$ which is guarantied to be positive. With $A\bm{\beta}$ as the forward gain, the forward function can now take scalar error $y-y'$ and produce vector $\mathbf{w'}$ as shown below.

\begin{align}
    \mathbf{w'} &= A\bm{\beta} \times (y-y') = A \times \bm{\beta} (y-y') \label{eq:error-func-formulation}
\end{align}

In (\ref{eq:error-func-formulation}), we can separate $\bm\beta$ from the forward gain and attach it to $y-y'$. This way we can keep using a forward gain of $A$ and use $\mathbf{e} = \bm\beta(y-y')$ as the new error. The error is now a function of scalar $y-y'$ which is shown by an error function block $E$ in Fig.~\ref{fig:system-setup}. We also notice that the error function is of the form $e = E(x) = ux$ as assumed in section \ref{sec:principle}. The error is calculated by multiplying the difference $y-y'$ with backward gain $\bm\beta$. Thus the gain of the error function is $\mathbf{u}=\bm\beta$.

\begin{figure}[!t]
\centering
\subfloat[$\sigma$=identity]{\includegraphics[width=0.23\textwidth]{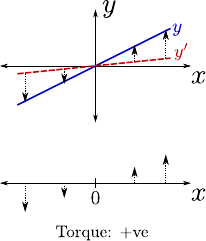}} \hfill
\subfloat[$\sigma$=ReLU]{\includegraphics[width=0.23\textwidth]{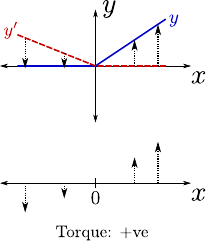}} \hfill
\subfloat[$\sigma$=tanh]{\includegraphics[width=0.23\textwidth]{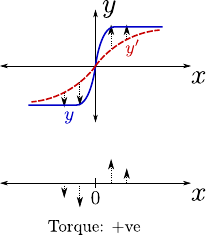}} \hfill
\subfloat[$\sigma$=sgn]{\includegraphics[width=0.23\textwidth]{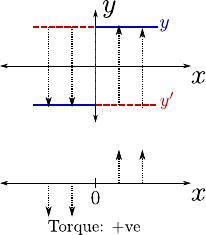}}
\caption{Torque analogy of error. The solid and dashed lines represent $y$ and $y'$ respectively. The difference $y-y'$ shown by the dotted arrows can be thought of forces acting on the x-axis. The resulting total torque $\sum_{k}(y^{[k]}-y'^{[k]})x^{[k]}$ is the output of the error function given by Eq.~(\ref{eq:multi-var-error}).}
\label{fig:torque}
\end{figure}

\section{Application in Machine Learning}\label{sec:application-in-ml}
In the following sections, we apply this method starting with simpler regression problems and then gradually develop methods for complex problems such as deep neural networks. 

\subsection{Regression on Noise-Free Data}
In machine learning, the activation functions can be unity, ReLU, tanh etc. The backward gain of the feedback system for any activation function is $\bm\beta={\partial y'}/{\partial\mathbf{w}}=\sigma'\mathbf{x}$ where $\sigma'$ is the gain of the activation function. From section~\ref{sec:stability-criteria}, we know that the gain of the error function needs to be $\mathbf{u}=\bm\beta$ in order to make the forward and backward gain products positive. We notice that if the activation function is monotonic and has non-negative $\sigma'$ then the source of negative component in $\bm\beta$ is only $\mathbf{x}$. Hence, we can simplify the gain of error function as $\mathbf{u}=\mathbf{x}$. For a single training sample, the error corresponding to $i^{th}$ weight is $e_{w_i} = u_i(y-y') = x_i(y-y')$. With many training samples the error is the sum of the errors from all the samples. The error for all the weights can be expressed as matrix multiplication, as in Eq.~(\ref{eq:multi-var-error}), where $u_{w_i}^{[k]} = x_i^{[k]}$ is the error gain for $i^{th}$ weight and $k^{th}$ sample. For all the training samples the error function gain becomes a matrix $\mathbf{U}$.
\begin{align}
     \mathbf{e} &= E(y-y') = \mathbf{U}(\mathbf{y}-\mathbf{y'})^T = \begin{bmatrix}u_{w_1}^{[1]} & u_{w_1}^{[2]} & \hdots & u_{w_1}^{[m]} \\
                                  u_{w_2}^{[1]} & u_{w_2}^{[2]} & \hdots & u_{w_2}^{[m]} \\
                                    \vdots  & \vdots    & \ddots & \vdots \\
                                  u_{w_n}^{[1]} & u_{w_n}^{[2]} & \hdots & u_{w_n}^{[m]} \\
                                  \end{bmatrix}
                    \begin{bmatrix}
                    y^{[1]} - y'^{[1]} \\ y^{[2]} - y'^{[2]} \\ \vdots \\ y^{[m]} - y'^{[m]}
                    \end{bmatrix}
                  =  \begin{bmatrix}
                    e_{w_1} \\ e_{w_2} \\ \vdots \\ e_{w_n}
                    \end{bmatrix}
                    \label{eq:multi-var-error}
\end{align}

It is interesting to see that no information of the activation function is needed in Eq.~(\ref{eq:multi-var-error}) in order to determine the weights. We provide an intuition for this using the analogy of torque in Fig.~\ref{fig:torque}. We choose a simple target data set $y$ which comes from a process $y=\sigma(wx)$ where weight $w$ is positive. Four different activation functions are shown in Fig.~\ref{fig:torque} with the training data $y$ (solid blue), $y'$ (dashed red) for randomly initialized weight. The error at each training sample is $(y-y')x$. The difference $y-y'$ (dotted arrow line) can be thought of as a force acting on the $x$ axis. Thus, there is a torque on the $x$ axis with origin as the axis of rotation. The sum of the torques represents the total error. In all four cases the resulting torque is positive. Thus, the weight will change in the positive direction. This is true regardless of the activation function being used as long as the gradient of the activation is non-negative. This makes it possible to perform regression even on the signum function $sgn$ which is non-differentiable. The $sgn$ function has zero gradient and also has a discontinuity at the origin. However, the gradient at the origin can be represented as a Dirac delta function which is infinite in value but is non-negative. The ReLU and tanh also have a non-negative gradient. Thus $\mathbf{u}=\mathbf{x}$ can be used in these cases. If the gradient of activation is negative then $\mathbf{u}=sgn(\sigma')\mathbf{x}$ is used. 

\begin{figure}[!t]
\centering
\subfloat[]{\includegraphics[width=0.3\textwidth]{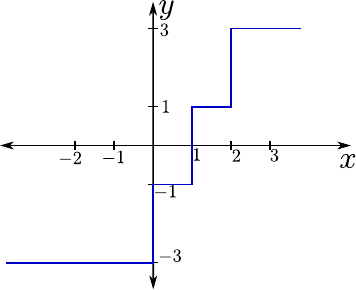} \label{fig:non-diff-func} } \hfill
\subfloat[]{\includegraphics[width=0.35\textwidth]{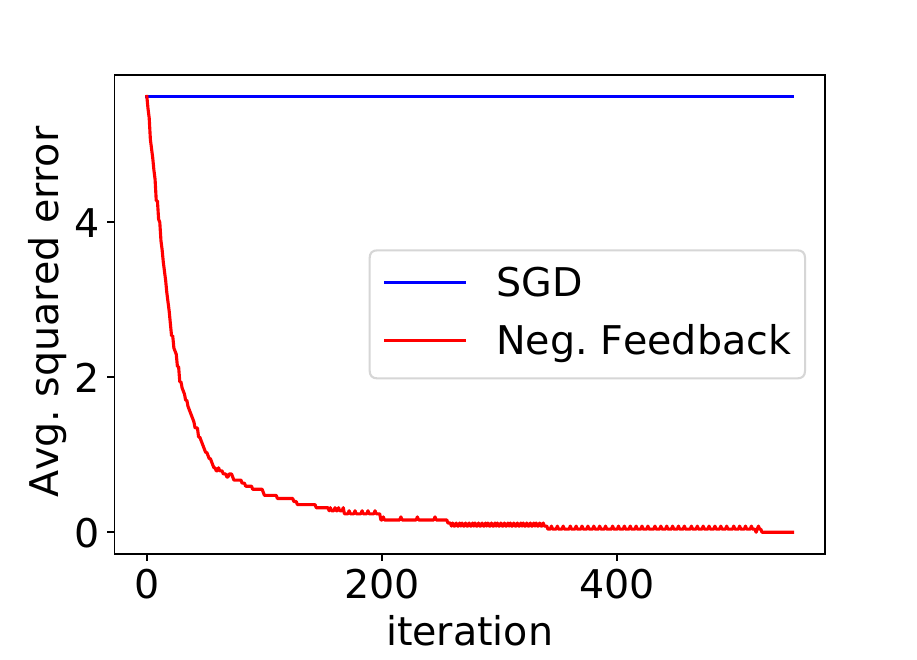}\label{fig:non-diff-vs-sgd} } \hfill
\subfloat[]{\includegraphics[width=0.3\textwidth]{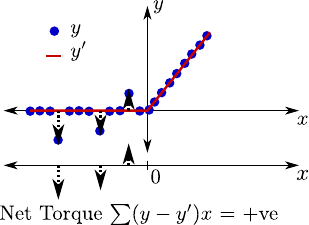} \label{fig:min-sq-1}}
\caption{(a) Target data $y=\sigma(wx+b)$ for $\sigma = sgn(z+1)+sgn(z)+sgn(z-1)$ where $z=wx+b$. The plot is shown for $w=1, b=-1$ (b) Average squared error during training for negative feedback system and SGD w.r.t training iteration. (c) Illustrating the failure of squared error minimization.}

\end{figure}

A comparison between training with negative feedback and Stochastic Gradient Descent (SGD) is shown in Fig.~\ref{fig:non-diff-vs-sgd} for a non-differentiable function in Fig.~\ref{fig:non-diff-func}. A value of $\gamma=1$ and $\eta=10^{-2}$ is used in Eq.~(\ref{eq:w-update-formulation}). For SGD, the gradient of the activation is used which is mostly zero. As a result average squared error in SGD does not change. However, the error is decreasing in the negative feedback optimization. It can correctly learn the weight and the bias terms without using the gradient of the activation

\subsection{Regression on Noisy Data}\label{sec:noisy-data}
Although the negative feedback system can learn a function properly with $\mathbf{u} = \mathbf{x}$, it can only do that with clean noise-free training data. For noisy training data, $\mathbf{u} = \mathbf{x}$ cannot achieve minimum squared error when a non-linear activation is present. The reason for this is shown in Fig.~\ref{fig:min-sq-1} using a ReLU activation. In the figure, although $y'$ and $y$ matches exactly on the linear part, the noisy data points on the saturated part of the activation (where ReLU is zero), still creates a positive torque. As a result, the negative feedback system optimizer will increase the slope further. In order to prevent that, we must isolate and ignore $y-y'$ for the saturated parts of the activation function by using a window function. The actual error function gain  $\mathbf{u} = \bm{\beta} = \sigma'\mathbf{x}$ gives us that window function as $\sigma'$. Thus when the torque $(y-y')x$ is multiplied with the window function $\sigma'$, the torque contribution from the saturated parts towards the net torque will be zero. With this interpretation of $\sigma'$ as the window function that isolates saturated and non-saturated parts we can approximate a window function for the non-differentiable functions as well and minimize squared error which we show in section~\ref{sec:window-func}.

\subsection{Single Layer Classifier}
The regression problem can be turned into a perceptron classifier by using softmax or tanh as the activation function. Hence, Eq.~(\ref{eq:multi-var-error}) also represents the error function for a single layer perceptorn.
For a multi class classifier the error function is simply the extension of Eq.~(\ref{eq:multi-var-error}). The single row of $\mathbf{y}-\mathbf{y}'$ becomes a matrix with $y-y'$ of different classes stacked as rows. 

\begin{figure}[!t]
\centering
\includegraphics[width=5.0in]{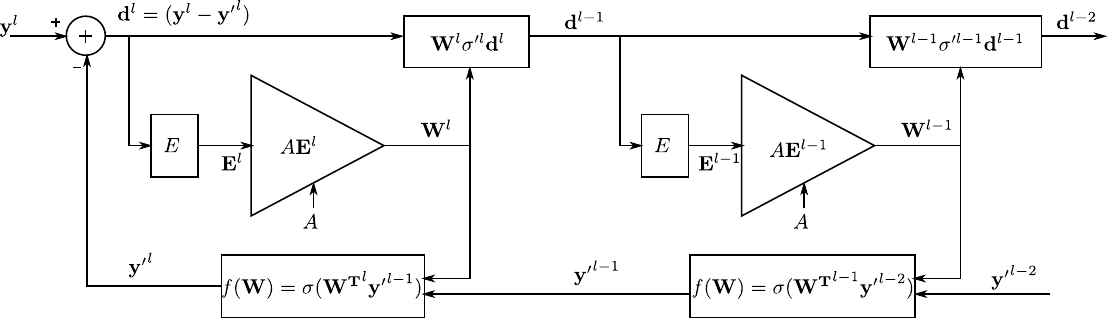}
\caption{Backpropagating the difference vector to previous layers.}
\label{fig:backprop}
\end{figure}

\subsection{Deep Network}
To use this system in deep networks, a method for error backpropagation is needed. A network is shown in Fig.~\ref{fig:backprop} with $l$ denoting layer number. The low pass filters haven been omitted in the figure for simplicity. The input from second to last layer $\mathbf{y'}^{l-2}$ generates the final output $\mathbf{y'}^l$. We treat the output $\mathbf{y'}^l$ as a result of the input $\mathbf{y'}^{l-2}$ as in Eq.~(\ref{eq:backprop-back-func}). For $c^{th}$ class output it can be expressed as Eq.~(\ref{eq:backprop-simple}). The backward gain for a weight is given by Eq.~(\ref{eq:backprop-back-gain}). Multiplying the difference $d^l_c = (y^l_c-y'^l_c)$ with the backward gain, we can write the error for $e_{w^{l-1}_{ji}}$ as Eq.~ (\ref{eq:backprop-error}). 
\begin{align}
    \mathbf{y'}^l &= \sigma(\mathbf{W^T}^l\sigma(\mathbf{W^T}^{l-1}\mathbf{y'}^{l-2})) \label{eq:backprop-back-func} \\
    y^l_c & = \sigma(\sum_iw^l_{ic}(\sigma(\sum_jw^{l-1}_{ji}y^{l-2}_j))_i) \label{eq:backprop-simple} \\
    \beta_{w^{l-1}_{ji,c}} &= \sigma'^{l}w^l_{ic}\sigma'^{l-1}y^{l-2}_j \label{eq:backprop-back-gain} \\
    e_{w^{l-1}_{ji}} & = \sigma'^{l-1}y^{l-2}_j \sum_c \sigma'^{l}w^{l}_{ic}d^{l}_c \label{eq:backprop-error} \\
    e_{w^{l-1}_{ji}} & = \sigma'^{l-1}y^{l-2}_j d^{l-1}_i = u_j^{l-1}d^{l-1}_i \label{eq:backprop-error-simple}
\end{align}
The sum over $c$ expresses the fact that every class output is influenced by $w^{l-1}_{ji}$. With $d^{l-1}_i=\sum_c \sigma'^{l}w^{l}_{ic}d^{l}_c$ in Eq.~(\ref{eq:backprop-error-simple}), $d^{l-1}_i$ can be thought of as the difference error for layer $l-1$. Also, $u_j^{l-1}$ represents the error function gain. The outcome is shown in Fig.~\ref{fig:backprop}. The difference vector of the last layer is multiplied with $\sigma'^{l}\mathbf{W}^l$ which produces the difference vector for the previous layer. This way error is backpropagated to all the previous layers.



\section{Comparison with Gradient Descent}\label{sec:comparison-grad}
For a negative feedback system it is important that the forward and backward gain product for each weight is positive. The gain of the error function as $\mathbf{u} = \bm{\beta}$ satisfies that condition. This condition is also necessary for squared error minimization as shown in section~\ref{sec:noisy-data}. In fact we can use $\mathbf{u} = \bm{\beta}^n$ as the gain as well where $n$ is an odd positive integer. This way the negative feedback system represents an infinite number of optimizers. The reason for odd positive $n$ is that it preserves the sign of $\bm\beta$. When $n=1$, the negative feedback system error implements the error gradient of the gradient descent optimization method. The gradient descent method minimizes a loss function, e.g. squared error as in Eq.~(\ref{eq:sq-err}). The weight parameters are updated by going in the opposite direction of the gradient which is given by Eq.~(\ref{eq:err-using-gradient}) for a weight $w_i$. Using $\mathbf{u} = \bm{\beta}$ in Eq.~(\ref{eq:multi-var-error}), the feedback error for a weight $w_i$ is given by Eq.~(\ref{eq:err-using-beta}). We see that both expressions are same except for a factor of $2/m$. The relationship between the two is $e_{w_i} = (m/2)(-\nabla q_{w_i})$. In gradient descent with a weight decay factor $\lambda$, the update rule is given by $w^t = w^{t-1} - \eta(\nabla q_{w_i} + \lambda w^{t-1})$. If we let $\eta \gets \eta\lambda$, $\gamma\gets2/(Am\lambda)$ and substitute $e_{w_i} = (m/2)(-\nabla q_{w_i})$ in Eq.~(\ref{eq:w-update-formulation}) we get Eq.~(\ref{eq:grad-update-rule}) which is exactly the same as gradient descent update rule.

\begin{align}
    q &= \frac{1}{m}\sum_k(y^{[k]} - y'^{[k]})^2 \label{eq:sq-err}\\
    -\nabla q_{w_i} &= \frac{2}{m}\sum_k(y^{[k]} - y'^{[k]}).\sigma'.x_i \label{eq:err-using-gradient} \\
    e_{w_i} &= \sum_k(y^{[k]}-y'^{[k]}).\sigma'.x_i \label{eq:err-using-beta} \\
    {w}_i^{t} &= {w}_i^{t-1} -\eta(\nabla q_{w_i} + \lambda {w}_i^{t-1}) \label{eq:grad-update-rule}
\end{align}

At this stage we can see that with $\mathbf{u}=\bm{\beta}$ which is the condition for squared error minimization, the negative feedback system and gradient descent method are equivalent. Also, by noticing Fig.~\ref{fig:backprop}, one can easily realize that the error propagation to previous layers is the same as the backpropagation technique in gradient descent method~\cite{Rumelhart1986}. We have derived it only using the properties of the negative feedback system. Thus, the negative feedback system allows us to look at and analyze the optimization problem from a different perspective. In gradient descent the objective is to minimize a loss function. However, in negative feedback system, the objective is inverse the backward function. 

\section{Optimization Techniques derived from Negative Feedback System}\label{sec:optimization}
In this section we will describe some commonly used optimization techniques used under gradient descent method which can be established from the properties of the negative feedback system. 

\subsection{Weight update with Adaptive Momentum}\label{sec:adal}

\begin{figure}[h]
\centering
\subfloat[]{ \includegraphics[width=2.3in]{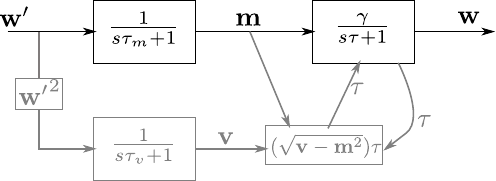} \label{fig:momentum}} \hfill
\subfloat[]{ \includegraphics[width=3.0in]{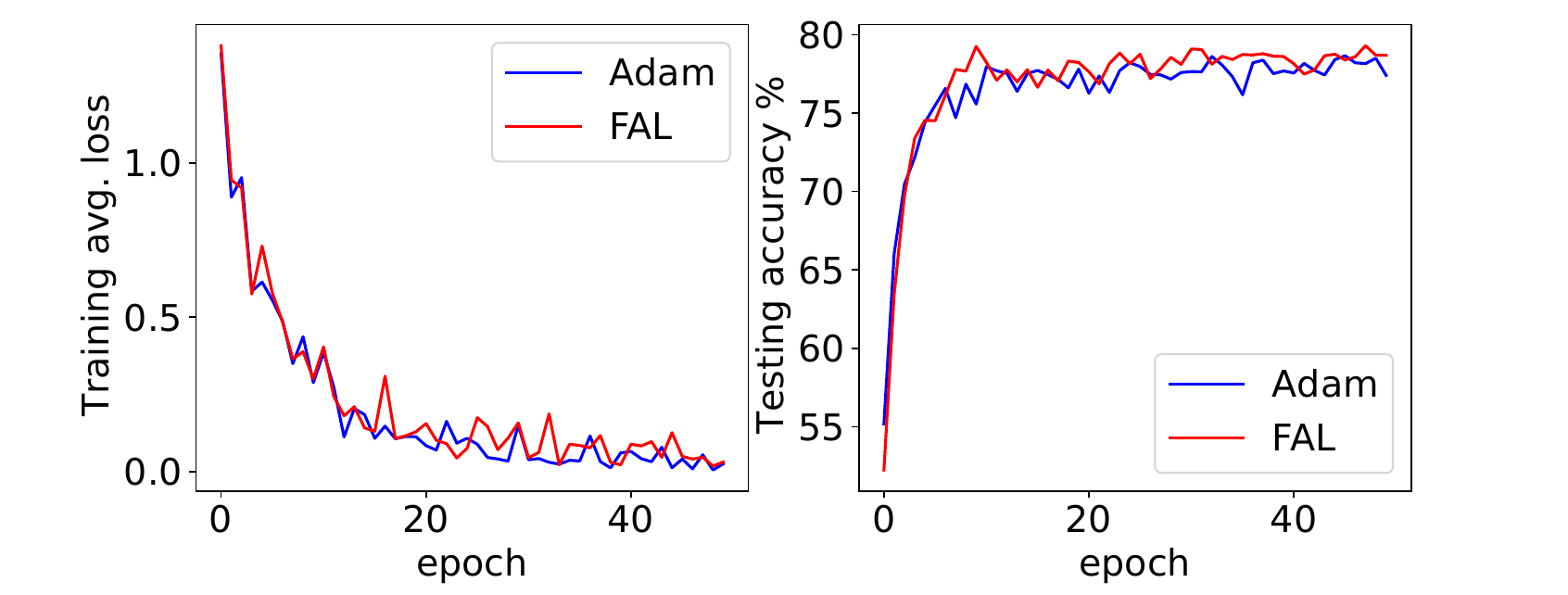} \label{fig:adam-vs-adal}}
\caption{(a) Two first order low pass filters in cascade implements weight update with momentum (dark lines). Another low pass filter (gray lines) implements Filter Assisted Learning rate (FAL). (b) Comparison between Adam and FAL on CIFAR10 using ResNet20.}
 
\end{figure}

The first order low pass filter in Fig.~\ref{fig:system-setup} may not be enough to control the speed in practice. In order to better control the stability, we can use a second order low pass filter which is just two first order low pass filters in cascade. This setup is shown in Fig.~\ref{fig:momentum} (dark lines). The output of the high gain forward function $\mathbf{w'}$ goes into the first low pass filter with time constant $\tau_m$. The output from that filter $\mathbf{m}$ goes into the second low pass filter with time constant $\tau$ which updates the weight. Similar to Eq.~(\ref{eq:w-update-formulation}), the output of the first low pass filter can be written as follows. 
\begin{align}
    \mathbf{m}^t &= \mathbf{m}^{t-1} + (\mathbf{w'} - \mathbf{m}^{t-1})\eta_m \quad 
    \text{( where $\eta_m=\frac{\partial t}{\tau_m}$)} \nonumber \\
    \mathbf{m}^t &= \beta_m\mathbf{m}^{t-1} + (1-\beta_m)\mathbf{w'} \quad
    \text{( where $\beta_m=1-\eta_m$)} \label{eq:momentum-formulation}
\end{align}

The expression in Eq.~(\ref{eq:momentum-formulation}) is exactly the same as the momentum estimation~\cite{Kingma2015AdamAM} as used in gradient descent. However, the filter system is capable of controlling the speed of the system only up to a certain value (given by the cut-off frequency of the filter). As a result, if the cut-off frequency is large (small $\tau_m$ or large $\eta_m$) substantial oscillations from $\mathbf{w'}$ can pass through. In order to stop the oscillation to pass through the second filter we need to increase $\tau$ (consequently decrease $\eta$). This can be done by detecting the presence of substantial oscillation in $\mathbf{w'}$ and decreasing $\eta$ proportionally by the strength of that oscillation. The strength of oscillation is proportional to the standard deviation of $\mathbf{w'}$ (the more $\mathbf{w'}$ deviates from the mean, the more the oscillation). The variance of $\mathbf{w'}$ is $Var(\mathbf{w'}) = E[\mathbf{w'}^2] - E[\mathbf{w'}]^2$ where $E[.]$ is expectation or average value. A first order low pass filter is in fact a moving average calculator. Hence, we already have $\mathbf{m}=E[\mathbf{w'}]$. Additionally, we can measure $\mathbf{v}=E[\mathbf{w'}^2]$ with another low pass filter with time constant $\tau_v$ as shown in Fig.~\ref{fig:momentum} (gray lines). Thus, $\tau$ can be proportionally scaled using the standard deviation $\sqrt{\mathbf{v}-\mathbf{m}^2}$. Using Eq.~(\ref{eq:momentum-formulation}), $\mathbf{v}$ is given by $\mathbf{v}^t = \beta_v\mathbf{v}^{t-1} + (1-\beta_v)\mathbf{w'}^2$. At this point, the stabilization method becomes extremely similar to the Adaptive Momentum (Adam) estimation~\cite{Kingma2015AdamAM} method. In Adam, the subtraction by $\mathbf{m}^2$ is not applied and there are bias correction process. A comparison between the performance between Adam and Filter Assisted Learning rate (FAL) derived from negative feedback system is shown in Fig.~\ref{fig:adam-vs-adal} with CIFAR10~\cite{cifar} dataset on ResNet20~\cite{He2016}. A learning rate of $10^{-3}$, $\lambda=1$ and a batch size of $128$ is used. For Adam default settings are used~\cite{Kingma2015AdamAM}. For FAL we have used $\beta_m = \beta_v = 0.99$ because both filters needs to be the same type of average calculator. Even without the bias correction in FAL, the performance is comparable. 

\subsection{Predicting and Fixing a Dead network}
The output of the error function is $\mathbf{e} = \bm{\beta}(\mathbf{y}-\mathbf{y'})$. The backward function inversion or optimization is complete when $\mathbf{y}=\mathbf{y'}$ and consequently $\mathbf{e}=0$. However, $\mathbf{e}$ can become zero even for $\mathbf{y}\neq\mathbf{y'}$ when $\bm{\beta}$ is zero. Hence, it is important to make sure that $\bm{\beta}\neq 0$ for negative feedback to work. Since, $\bm{\beta}=\sigma'\mathbf{x}$, we need to make sure that either $\sigma'$ or $\mathbf{x}$ (inputs to a layer) or both is not zero for all the training samples. With this critrion, it is now easy to predict a dead network. In Fig.~\ref{fig:backprop}, if layer inputs $\mathbf{y'}^{l-1}$ and $\mathbf{y'}^{l-2}$ become zero, negative feedback system stops working as an optimizer. This is the same phenomenon as seen in the dying ReLU~\cite{Lu2020DyingRelu} problem. A straightforward fix would be to use an activation function that does not saturate to zero. A leaky ReLU~\cite{He2015ReLU} or exponential ReLU~\cite{DjorkELU} naturally satisfies this condition. 

\subsection{Fixing Vanishing Gradient}
\begin{figure}[!t]
\centering
\subfloat[]{ \includegraphics[width=0.3\textwidth]{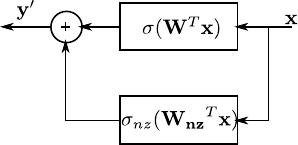}  \label{fig:fix-vanish-grad}} \hfill
\subfloat[]{ \includegraphics[width=0.5\textwidth]{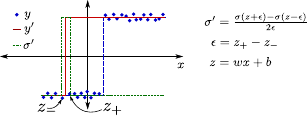}  \label{fig:approx-window-func}}
  \caption{(a) Fixing the vanishing gradient problem by introducing non-zero gradient term in the backward path. (b) Approximating the gradient of non-differentiable function.}
 
\end{figure}

As mentioned in the previous paragraph, negative feedback stops working as an optimizer when either $\sigma'$ or inputs to a layer becomes zero. In the case when $\sigma'$ is zero or close to zero, a similar phenomenon happens but this time because of the gradient of the activation function called the vanishing gradient problem. From negative feedback system analysis, the solution is to make sure that $\bm\beta\neq 0$. A solution would be to use leaky ReLU as activation function. Another solution would be to introduce a non-zero additive gradient term $\sigma'_{nz}$ in the backward gain expression as $\bm\beta = (\sigma' + \sigma'_{nz})\mathbf{x}$. The resulting backward path function takes the form as shown in Fig.~\ref{fig:fix-vanish-grad}. If we choose $\sigma_{nz}$ as an identity function, we get the skip connection of the residual network as described in \cite{He2016}. The matrix $W_{nz}$ matches the size of $\mathbf{x}$ with the size of $\mathbf{y'}$. When the size of $\mathbf{y}$ and $\mathbf{x}$ are same $W_{nz}$ can be replaced by an identity mapping. In \cite{He2016}, the authors hypothesize that it is easier to train a deep network by skip connection. From the framework of negative feedback system, the idea of the skip connection comes naturally when we try to make the backward gain $\bm\beta$ non-zero.

\subsection{Window Function for Non-Differentiable Activation}\label{sec:window-func}
In section~\ref{sec:noisy-data} we have shown that a window function is necessary in order to minimize squared error. The window function sets $y-y'$ to zero where activation is saturated and accounts for $y-y'$ where activation is changing. For differentiable functions, the gradient of the activation $\sigma'$ can serve as the window function. For non-differentiable activation functions we can use an approximated window function as shown in Fig.~\ref{fig:approx-window-func}. A signum function is used an example. A window function is taken as the approximated gradient while making sure that it captures the changing part of the activation. This can be done using the two points immediately next to the discontinuity ($z_+$ and $z_-$) as shown in Fig.~\ref{fig:approx-window-func}. For stability purpose, in order to avoid very large gradients, the resulting gradient is clipped to a value of 1. However, the process of finding the points $z_+$ and $z_-$ can be computationally intensive specially when the input data is not sorted. 

\subsection{Weight Transport}
It is widely accepted that backpropagation is not compatible with the biological brain because there are no known mechanisms how the synaptic weights are transmitted to the backpropagation network~\cite{Qilani2016, Lillicrap2016, Will2018}. However, Fig.~\ref{fig:backprop} might shed some light on how the weight transport can happen. Instead of transmitting the synaptic weight from the backward path (`backward' in terms of the negative feedback topology) to the backpropagation path, both the paths draw the weights from the high forward gain block. The high forward gain block acts as the weight generator which supplies the weights to backward path function and backpropagation path function. The forward gain block is not a matrix multiplying block. However, together with the error function block the forward gain block can be considered as a matrix multiplying block. Thus functionally it is the same type of block as the backward path blocks and the backpropagation path blocks. Hence, it can be implemented by a neuron as well. Thus, we can hypothesize that there could be neurons in the brain that generate the weight-signals by which the other neurons adjust their respective weighs. 

\section{Discussion}
Although, the negative feedback system seems to offer some solutions to dying ReLU and vanishing gradient problem, it does not seem to offer any solution to the weight initialization method. If the initial weight is negative for the particular case in Fig.~\ref{fig:approx-window-func}, the training will fail. However, deep networks can still be trained if weights are initialized properly~\cite{xavier-init, He2015ReLU, Lu2020DyingRelu}. Weight initialization depends on the statistical properties of the network and data. However, in the negative feedback system, the stability constraint $\mathbf{u}=\sigma'\mathbf{x}$ is not concerned with the statistics of the data. For the same reason, it also does not predict the batch normalization~\cite{batch-norm} process which have been proven to be useful in training deep networks. In this paper we have mostly focused our discussion on the neural network optimization. The generalization to other optimization methods such as particle swarm~\cite{particle-swarm}, genetic algorithm~\cite{Sun2020} or neuroevolution\cite{Stanley2019} needs further investigation.

\section{Conclusion}
In this paper we have presented the negative feedback system as the physical process of computation for optimization. We think it can connect learning algorithm and analog circuit design thereby paving the way for low power optimization circuits by utilizing the physics of electronic devices~\cite{Indiveri2011, Qiao2015, Sarpeshkar1998}. Many optimization techniques are easily derived from the properties of negative feedback system such as weight decay and adaptive momentum. The negative feedback system also predicts the problems associated with training deep network such as dying ReLU and vanishing gradient. Not only that it also offers solution to those problems in an explainable way. We believe this novel view of neural network optimization is capable of providing valuable insights on the optimization techniques which many open problems in machine learning can benefit from.








\bibliographystyle{plainnat}
\bibliography{bibfile}

\end{document}